\documentclass[wcp]{jmlr}

\usepackage[T1]{fontenc}
\usepackage{longtable}

\usepackage{multirow}
\usepackage{booktabs}
\usepackage{float}
\usepackage{placeins}

\usepackage{lineno}

\makeatletter
\let\Ginclude@graphics\@org@Ginclude@graphics 
\makeatother

\jmlryear{2026}

\title{NIR: Approximate Neural Instance Retrieval for $\mathcal{ALCIQ}$}

 \author{\Name{N'Dah Jean Kouagou} \Email{ndah.jean.kouagou@upb.de}\\
  \Name{Pritilata Saha} \Email{psaha@mail.uni-paderborn.de}\\
  \Name{Caglar Demir} \Email{caglar.demir@upb.de}\\
  \Name{Axel-Cyrille Ngonga Ngomo} \Email{axel.ngonga@upb.de}\\
\addr Data Science Group, Heinz Nixdorf Institute, Paderborn University, Germany}

\usepackage{pgfplots}
\pgfplotsset{
    compat=1.18,
    every axis/.append style={
        tick label style={font=\scriptsize},
        label style={font=\small\bfseries},
    },
}
\usepgfplotslibrary{groupplots}
\usetikzlibrary{calc}

\usepackage{tikz}
\usepackage{pgfplots}
\usepackage{pgfplotstable}
\usepackage{comment}

\begin{document}

\newcommand{\dl}{$\mathcal{ALCIQ}$\xspace}
\newcommand{\nand}{\ensuremath{\mathbf{\textsc{nand}}}\xspace}
\newcommand{\emb}{\ensuremath{\mathbf}}
\newcommand{\embi}[1]{\ensuremath{\mathbf{e}_{#1}}}
\newcommand{\Quaternions}{\ensuremath{\mathbb{H}}\xspace}
\newcommand{\real}{\mathrm{Re}}
\newcommand{\ConvolutionWithProjection}{\text{conv}}
\newcommand{\an}[1]{\todo[inline,color=cyan]{AN: #1}}
\newcommand{\ii}{\ensuremath{\mathcal{I}}}

\makeatletter
\let \@jmlrpages \@empty
\makeatother

\maketitle

\begin{abstract}
    Instance retrieval is the main bottleneck of concept-learning systems based on refinement operators. We present \emph{NIR}, a family of neural instance retrievers for $\mathcal{ALCIQ}$ that output graded membership scores from knowledge graph embeddings. We compare tokenization-based encoders (GRU, LSTM, Transformer) with a Compositional encoder that mirrors DL operators via primitive sub-networks. On eight benchmarks, all variants approximate Pellet closely on complete knowledge bases, while the Transformer model is most robust under incompleteness. The source code of our approach is at \url{https://github.com/dice-group/owlapy}.
\end{abstract}

\section{Introduction}
Class expression learning (CEL)~\cite{lehmann2011class,kouagou2025fast}, also known as concept learning, allows for explainable node classification on RDF knowledge graphs (KGs). Given a set of positive and a set of negative examples, CEL approaches learn a concept in a description logic (DL) that describes (ideally all) positive examples while excluding as many negative examples as possible~\cite{demir2023neuro,kouagou2025fast}. For instance, $\geq 4\,\texttt{numWheels}.\top\ \sqcap\ \exists\,\texttt{supports}.\texttt{WiFi}\ \sqcap\ \exists\,\texttt{canMove}.\texttt{Object}$ is a concept in the DL $\mathcal{ALCQ}$ which describes the skills of machines in Industry 4.0~\cite{demir2022rapid}. Individuals (e.g., machines) are classified as \emph{positive} if they have at least four wheels, support WiFi, and can move at least one object. Importantly, these skills are directly encoded into the syntactic representation of the class expression and are both human-readable (e.g., $\geq 4\,\texttt{numWheels}.\top$, $\exists\,\texttt{supports}.\texttt{WiFi}$) and machine-readable (see Table~\ref{tab:dl} for semantics of description logics). Consequently, concepts are white-box models~\cite{kouagou2025fast} and hence well suited for the implementation of explainable AI pipelines.

Most existing approaches for CEL are search-based and rely on symbolic reasoners (e.g., Pellet~\cite{sirin2007pellet} and HermiT~\cite{glimm2014hermit}) to retrieve instances of candidate concepts and to subsequently evaluate the quality of these concepts. Instance retrieval is known as the main bottleneck of CEL~\cite{karalis2024evaluating,sarker2019efficient}.  
This is because a reasoner is invoked repeatedly to retrieve instances of class expressions during CEL\textemdash typically, thousands of invocations are required to solve a learning problem. Another inherent limitation of symbolic reasoners is their inability to deal with incompleteness in data~\cite{DBLP:conf/kcap/TeyouFKD0HN25}\textemdash they lack native support for approximate reasoning when assertions are missing.

During CEL, each candidate concept must therefore be evaluated by deciding, for many individuals, whether they are instances of that concept. To make this concrete, consider the Animals benchmark from DL-Learner, one of the eight knowledge bases used in our evaluation. Its ABox contains 20 atomic assertions such as \texttt{Eel(eel01)} and \texttt{Eagle(eagle01)}; names like \texttt{animals\#eel01} refer to individual \texttt{eel01} in the benchmark namespace. The TBox defines taxonomic and property constraints, e.g.,
\[
\texttt{Eel} \sqsubseteq \texttt{Animal} \sqcap \texttt{HasEggs} \sqcap \texttt{HasGills} \sqcap \exists\,\texttt{habitat}.\texttt{Water}\, \sqcap ={0}\,\texttt{hasBodyPart}.\texttt{Leg}.
\]
A CEL system may ask whether \texttt{eagle01} is an instance of a complex candidate concept such as \texttt{$\neg$Penguin $\sqcap$ $\forall$hasCovering.Feathers}. A symbolic retriever answers from what is entailed by the available axioms under standard DL semantics. When ABox assertions are randomly removed to simulate incompleteness (Section~\ref{sec:experiments}), exact retrieval can miss individuals whose supporting evidence is absent, whereas knowledge graph embeddings can encode regularities across individuals and relations that remain informative under missing assertions.

Several approaches aim to reduce the overall cost of CEL, including pruning~\cite{lehmann2011class}, neural class expression synthesizers (NCES)~\cite{kouagou2023neural}, ECII~\cite{sarker2019efficient}, and semantics-aware caching~\cite{teyou2026semanticsawarecachingconceptlearning}. NCES use neural networks to directly synthesize solutions to learning problems without a search process, but require pretraining data that is itself constructed with symbolic reasoners. ECII invokes a reasoner only once to materialize an input knowledge base, then constructs candidate solutions from role fillers and Boolean operators, yet with limited expressiveness. None of these approaches replace the repeated instance retrieval step with a fast, embedding-based retriever that remains effective on incomplete knowledge bases.

In this work, we investigate neural networks and knowledge graph embeddings for approximate, scalable instance retrieval. We introduce $\mathrm{NIR}$, a family of retrievers for \dl\ that estimate $p(C(a))$
given a concept $C$ and an individual $a$. Rather than committing to a single architecture, we compare tokenization-based models (GRU, LSTM, Transformer), which read the syntax of $C$ as a token sequence, with a Compositional model that encodes DL operators through dedicated sub-networks. Our evaluation on eight datasets shows that all variants achieve high agreement with Pellet on complete knowledge bases, but they differ in speed, robustness to incompleteness, and extensibility.

The rest of the paper is organized as follows. Section~\ref{sec:background} presents the necessary background for understanding the paper. In Section~\ref{sec:related-work}, we describe related works and highlight their strengths and weaknesses. In Section~\ref{sec:approach}, we present our approach for robust instance retrieval, and experimental results in Section~\ref{sec:experiments}. Section~\ref{sec:conclusion} concludes the paper.  

\section{Background}
\label{sec:background}
\subsection{Description Logics and Reasoning}
Description logics are a family of languages for formal representation of knowledge~\cite{baader2003description,kouagou2025fast}. They have applications in artificial intelligence, the semantic web, bio-informatics, and automated reasoning~\cite{kouagou2025fast}.
In this work, we target the description logic \dl which extends the well-known and widely adopted $\mathcal{A}$ttributive $\mathcal{L}$anguage with $\mathcal{C}$omplements ($\mathcal{ALC}$) by adding qualified number restrictions and inverse roles. Let $A$ be a named concept, $R$ a named role, and $n$ a natural number with $n \geq 0$. Concepts in \dl can be built as follows:
\begin{enumerate}
    \item  $\top, \bot, A$ are concepts,
    \item If $R$ is an abstract role, then $R^-$ is also a role (inverse role),
    \item If $C, D$ are concepts and $R$ a role, then the following are concepts:
    $ C\ \sqcap\ D, C\ \sqcup\ D, \neg C, \forall\ R.C, \exists\ R.C, \geq{n}\ R.C, \leq{n}\ R.C, \text{ and } ={n}\ R.C $
\end{enumerate}

The syntax and semantics of \dl are presented in Table~\ref{tab:dl}. In the table, $a$ and $b$ are individuals, $\ii=(\Delta^\ii, \cdot^\ii)$ is an interpretation, where $\cdot^\ii$ is the interpretation function and $\Delta^\ii$ the domain.
\begin{table}[ht]
\centering
\caption{Syntax and Semantics of \dl. $\ii=(\Delta^\ii, \cdot^\ii)$ is an interpretation.}
\label{tab:dl}
\begin{tabular}{lcc}
\toprule
Construct & Synthax     & Semantics \\ \midrule
Top concept & $\top$      & $\Delta^\ii$ \\
Bottom concept & $\bot$      & $\emptyset$ \\
Atomic concept & $C$      & $\{a \in \Delta^\ii: a \in C^\ii\}$ \\
Negation & $\neg C$          & $\Delta^\ii \backslash C^\ii$ \\
Conjunction & $C \sqcap D$      & $C^\ii \cap D^\ii$ \\
Disjunction & $C \sqcup D$      & $C^\ii \cup D^\ii$ \\
Universal role restriction & $\forall\ R.C$     & $\{a: (a, b) \in R^\ii \rightarrow b \in C^\ii\}$ \\
Existential role restriction & $\exists\ R.C$     & $\{a: \exists\ b\in C \mbox{ and } (a, b) \in R^\ii\}$ \\
Min. cardinality restriction & $\geq{n}\ R.C$   & $\{a: |\{b: (a, b) \in R^\ii \wedge b \in C^\ii\}| \geq n\}$ \\
Max. cardinality restriction & $\leq{n}\ R.C$   & $\{a: |\{b: (a, b) \in R^\ii \wedge b \in C^\ii\}| \leq n\}$ \\
Exact cardinality restriction & $={n}\ R.C$   & $\{a: |\{b: (a, b) \in R^\ii \wedge b \in C^\ii\}| = n\}$ \\
\bottomrule
\end{tabular}
\end{table}
Reasoning in expressive DLs is computationally expensive~\cite{Kazakov08}.
Additionally, real-world DL knowledge bases (e.g., DBpedia, Wikidata) are large and often incomplete, prompting the need for more robust and scalable reasoning engines for practical applications. Existing
DL reasoners such as HermiT~\cite{glimm2014hermit} and Pellet~\cite{sirin2007pellet} implement hard-coded heuristics (e.g., the hyper-tableau algorithm) for reasoning. They cannot infer plausible assertions that are not directly entailable
by the considered knowledge base, which can be a shortcoming on incomplete knowledge bases. The approach we propose herein supports approximate reasoning and can inherently handle incomplete DL knowledge bases by leveraging knowledge graph embeddings.

\subsection{Knowledge Graph Embeddings}
Knowledge graphs are graph-structured knowledge bases 
intended to gather and convey knowledge of the real world, with nodes and edges representing entities and relation types, respectively~\cite{DBLP:journals/csur/HoganBCdMGKGNNN21}. To facilitate downstream applications, knowledge graphs are embedded into vector spaces such as $\mathbb{R}^d$~\cite{bordes2013translating}, $\mathbb{C}^d$~\cite{trouillon2016complex}, or even Clifford spaces $Cl_{p,q,r}$~\cite{kamdem2024embedding}. This is often achieved by using a knowledge graph embedding (KGE) model $\Phi$ to map entities and relation types to vectors such that most relationships within the original graph are retained. Most importantly, the KGE model allows for new links to be inferred using scoring functions, which is essential for knowledge completion~\cite{shen2022comprehensive}. 
KGEs also have applications in explainable AI. For instance, NCES~\cite{kouagou2023neural} and Drill~\cite{demir2023neuro} leverage knowledge graph embeddings to tackle CEL at scale. Our proposed approach employs KGEs the same way for robust and scalable instance retrieval, see Section~\ref{sec:approach}.

\section{Related Work}
\label{sec:related-work}
Several recent studies have investigated generating ontology embeddings~\cite{chen2025ontology}. 
However, most of them do not support instance retrieval queries or are limited only to atomic concepts.
Other works explored representing description logic concepts as geometrical shapes, e.g., boxes, roles as affine transformations, individuals as points, and then jointly training the resulting model~\cite{xiong2022faithful}. Yet, these works support only the lightweight DL $\mathcal{EL}++$, focus on learning concept inclusions, and are not extensively evaluated on instance retrieval.

For expressive DLs such as $\mathcal{ALC}$, approaches focusing primarily on ontology satisfiability have been proposed~\cite{ozccep2020cone}. 
The core aim of these approaches is to show that an ontology is satisfiable if and only if it admits a satisfying \emph{geometrical} structure. For instance, \cite{ozccep2020cone} proposed to use al-cones, i.e., sets $X$ of the form $X=\Pi_{i=1}^d X_i$ with $X_i\in \{\mathbb{R}, \mathbb{R}_+,\mathbb{R}_-, \{0\}\}$, to represent concepts in $\mathcal{ALC}$, but it is unclear how instances of these concepts can be retrieved.
Other works such as \cite{chen2021owl2vec} map OWL ontologies to RDF knowledge graphs
and employ random walks and Word2Vec~\cite{church2017word2vec} to learn class subsumptions, without specifying how instance retrieval can be performed.

To overcome the runtime limitations of instance retrieval engines, \cite{teyou2026semanticsawarecachingconceptlearning} proposed a semantics-aware caching algorithm. 
Their approach employs a subsumption-aware cache that maps concepts to instance sets using crisp set operations. ECII~\cite{sarker2019efficient} proposed to use an instance retrieval engine only once to fully materialize a given knowledge base before class expression learning. Afterwards, candidate class expressions for each learning problem are constructed by strategically applying role restrictions and conjunction/disjunction operators following membership information of input examples and their relationships to other individuals in the materialized knowledge base.
The embedding-based reasoner (EBR)~\cite{DBLP:conf/kcap/TeyouFKD0HN25} uses knowledge graph embedding models and description logic semantics to translate the instance retrieval problem into complex query answering over an input knowledge base. While this approach supports approximate reasoning as ours, it is computationally intensive on large datasets and often fails on long class expressions as suggested by our experimental results.  


\section{Proposed Methods}
\label{sec:approach}
We study two encoder designs for $\mathrm{NIR}$. \emph{Tokenization-based} models treat a concept as a token sequence and are our default choice in experiments when knowledge bases may be incomplete. The \emph{Compositional} model instead composes primitive networks according to the logical structure of $C$; it tracks Pellet closely on complete knowledge bases and can be extended with new sub-networks for additional constructors, but relies on Pellet to encode atomic concepts.

\subsection{Tokenization-based Approaches}
\label{sec:tokenization}
Tokenization-based models use the same configurable embedding dimension $d$ as the Compositional retriever; in our experiments, $d{=}256$, with hidden size $128$, max sequence length $128$, vocabulary size $512$, and dropout $0.1$. A concept is tokenized (byte-pair encoding~\cite{sennrich2015neural}) and mapped to token embeddings $\mathbf{x}_i \in \mathbb{R}^{d}$.
\subsubsection{RNN-based Approach.} Three GRU or LSTM layers~\cite{cho2014learning,hochreiter1997long} are followed by two linear layers. Given hidden states $\mathbf{h}_1,\ldots,\mathbf{h}_n$ and individual embedding $\mathbf{a}$,
\begin{align}
\mathbf{h} = \frac{1}{n} \sum_{i=1}^{n} \mathbf{h}_i,\ 
\mathbf{u} = \mathrm{concat}(\mathbf{h},\mathbf{a}),\
p = sigmoid(\mathrm{Lin}(\mathbf{u})),
\end{align}
where $\mathrm{Lin(.)}$ denotes linear layers with GELU activation, and $p$ the probability that individual $a$ is an instance of the given concept $C$.

\subsubsection{Transformer-based Approach.} 
A $3$-layer Transformer encoder~\cite{vaswani2017attention} with $4$ heads applies positional encoding to token embeddings, uses average pooling to obtain $\mathbf{v}$, and concatenates $\mathbf{v}$ with $\mathbf{a}$ before the same two linear layers as above.

\subsection{Compositional Neural Instance Retriever}
\label{sec:compositional}
The semantics of \dl can be reduced to a small set of primitives. The binary connective \nand~\cite{frank2004review}, defined by $\nand(A,B) := \neg (A \sqcap B)$, is functionally complete for the Boolean constructors: negation is $\neg C = \nand(C,C)$, union is $C \sqcup D = \nand(\neg C, \neg D)$, and intersection follows via De Morgan as $C \sqcap D = \neg(\neg C \sqcup \neg D)$. Together with role inversion and cardinality restrictions, we obtain a Compositional retriever that mirrors the syntax of \dl\ through primitive sub-networks. New constructors can be supported by adding further sub-networks without retraining the entire model.

We assume a KGE model that maps each individual $a$ and role $R$ to vectors $\mathbf{a}, \mathbf{R} \in \mathbb{R}^{d}$, where $d$ is the embedding dimension of the chosen KGE model and can be set to any value supported by that model (we use $d{=}256$ in our experiments). Table~\ref{tab:nir-notation} summarizes the notation used in this section. The encoder $\phi$ composes three primitive maps $\phi_\nand$, $\phi_\le$, and $\phi_{\mathrm{inv}}$ (GELU feed-forward, hidden size $128$); each is implemented as a neural module $(\phi_{\ldots},f_{\phi_{\ldots}})$---for example, $(\phi_\nand,f_{\phi_\nand})$---where $f_{\phi_{\ldots}}$ is the scoring head that returns a logit for $(C,a)$. Atomic concepts are mapped as $\phi(C)=h(\rho(C))$, where $\rho$ is Pellet~\cite{sirin2007pellet}, $h$ is PMA~\cite{lee2019set}, $\phi(\bot)=\emptyset$, and $\rho(\top)$ returns all individuals. Let $\phi_{\mathrm{last}} \in \{\phi_\nand,\phi_\le,\phi_{\mathrm{inv}}\}$ denote the outermost primitive in the parse tree of $C$. We then define $\mathrm{NIR}(\phi(C),\mathbf{a}) := \sigma(f_{\phi_{\mathrm{last}}}(\phi(C),\mathbf{a})) \in [0,1]$ as an estimate of $p(C(a))$.

\begin{table}[t]
\centering
\caption{Notation for the Compositional retriever (our experiments use $d{=}256$, hidden size $128$, dropout $0.1$, threshold $\gamma{=}0.5$).}
\label{tab:nir-notation}
\resizebox{\textwidth}{!}{
\begin{tabular}{@{}ll@{}}
\toprule
Symbol & Meaning \\
\midrule
$\phi(C)$ & embedding of concept $C$ (composition of $\phi_\nand$, $\phi_\le$, $\phi_{\mathrm{inv}}$) \\
$\mathbf{a}$, $\mathbf{R}$ & KGE vectors for individual $a$ and role $R$ \\
$\rho(C)$, $h(\cdot)$ & instance set from Pellet; PMA pooling over all $\mathbf{a}$'s from $C$ \\
$\phi_\nand,\phi_\le,\phi_{\mathrm{inv}}$ & primitive maps used in $\phi(\cdot)$ \\
$(\phi_\nand,f_{\phi_\nand})$, $(\phi_\le,f_{\phi_\le})$, $(\phi_{\mathrm{inv}},f_{\phi_{\mathrm{inv}}})$ & neural modules (map and scoring head) \\
$f_{\phi_{\ldots}}$ & scoring head of primitive $\phi_{\ldots}$ (logit for $(C,a)$) \\
$\phi_{\mathrm{last}}$ & outermost primitive in the parse tree of $C$ (one of $\phi_\nand$, $\phi_\le$, $\phi_{\mathrm{inv}}$) \\
\bottomrule
\end{tabular}
}
\end{table}

Complex concepts are mapped recursively by $\phi$; $\mathbf{R}$ denotes the KGE vector of role $R$. Table~\ref{tab:enc-rules} lists the non-Boolean cases; $\neg$, $\sqcup$, and $\sqcap$ follow from $\phi_\nand$ via the reductions at the start of this section.
\begin{table}[t]
\centering
\small
\setlength{\tabcolsep}{5pt}
\caption{Recursive definition of $\phi$.}
\label{tab:enc-rules}
\begin{tabular}{@{}l@{\;}l@{}}
\toprule
Constructor & $\phi(\cdot)$ \\
\midrule
$\exists\,R.C$ & $\phi_\nand\!\bigl(\phi_\le(0,\mathbf{R},\phi(C)), \phi_\le(0,\mathbf{R},\phi(C))\bigr)$ \\
$\forall\,R.C$ & $\phi_\le(0,\mathbf{R},\phi(\neg C))$ \\
$R^-$ & $\phi_{\mathrm{inv}}(\mathbf{R})$ \\
$\le n\,R.C$ & $\phi_\le(n,\mathbf{R},\phi(C))$ \\
$\ge n\,R.C$, $n\ge 1$ & $\phi_\nand\!\bigl(\phi_\le(n{-}1,\mathbf{R},\phi(C)), \phi_\le(n{-}1,\mathbf{R},\phi(C))\bigr)$ \\
$\ge 0\,R.C$ & $\phi(\top)$ \\
\bottomrule
\end{tabular}
\end{table}
Each row reduces an \dl\ constructor to the primitive maps $\phi_\nand$, $\phi_\le$, and $\phi_{\mathrm{inv}}$. Universal restrictions use $\forall R.C \equiv \neg\exists R.\neg C$, implemented as $\phi_\le(0,\mathbf{R},\phi(\neg C))$. Existential and at-least-$n$ restrictions are cardinality-based: $\exists R.C \equiv \ge 1\,R.C$, and for $n\ge 1$ we have $\ge n\,R.C \equiv \neg(\le n{-}1\,R.C)$; the case $n{=}0$ is $\phi(\top)$ since every individual is either $R$-linked to at least one individual of type $C$ or to exactly $0$ individuals of type $C$. Accordingly, $\phi(\exists R.C)$ and $\phi(\ge n\,R.C)$ apply $\phi_\nand$ to two copies of the same $\phi_\le$ output---the same pattern as $\phi(\neg C)=\phi_\nand(\phi(C),\phi(C))$, because each applies a negated $\le$-restriction. Role inversion uses $\phi_{\mathrm{inv}}$; $\le n\,R.C$ is handled directly by $\phi_\le$.
PMA is pretrained on atomic concepts using positive instances $E^+$ and sampled non-members $E^-$; we use $\gamma=0.5$ to separate them. Complex concept embeddings are then obtained by composing the rules in Tables~\ref{tab:nir-notation} and~\ref{tab:enc-rules}.

\subsubsection{Scoring Technique.}
Each neural module $(\phi_{\ldots},f_{\phi_{\ldots}})$ applies $\phi_{\ldots}$ to obtain a $d$-dimensional concept embedding. At inference, only $f_{\phi_{\mathrm{last}}}$ is used: $\mathrm{NIR}(\phi(C),\mathbf{a})=\sigma(f_{\phi_{\mathrm{last}}}(\phi(C),\mathbf{a}))$, where $\sigma$ is the \emph{sigmoid} activation function.

The main network directly integrates the \emph{symmetry} and \emph{monotonicity} properties of $\phi_\nand$ and $\phi_\le$, respectively. Specifically, the monotonicity of $\phi_\le$ (i.e., the concept $\le n\ R.C$ is always subsumed by $\le (n+1)\ R.C$) is ensured by adding $sigmoid(n)$ to output logits before the final probability score is computed for a given individual. The symmetry of $\phi_\nand$ is obtained by summing the hidden representations of its arguments.

Below, only the maps $\phi_{\ldots}$ are used, since we do not score an individual yet.
\begin{example}
\label{ex:encoding}
 $\neg(\leq 2 \texttt{hasSuperStar.Footballer}) \sqcap \texttt{Club}$.
    \label{eq:Fifaexampleconcept}
\end{example}
The embedding of Example~\ref{eq:Fifaexampleconcept} (\emph{a club with at least two superstar footballers}) is computed by $\phi$ as follows.
\begin{enumerate}
    \item [a.] $\leq 2\ $ \texttt{hasSuperStar.Footballer}:
     $\emb{x} = \phi_\le(2, \mathbf{R}_{\texttt{hasSuperStar}}, \phi(\texttt{Footballer})).$  

    \item [b.] $\neg(\leq 2 \ \texttt{hasSuperStar.Footballer})$:
    $\emb{x}' = \phi_\nand(\emb{x}, \emb{x}).$
    
    \item [c.] $\emb{y} = \phi(\texttt{Club})$.
    
    \item [d.] $\emb{e} = \phi_\nand(\phi_\nand(\emb{x}', \emb{y}), \phi_\nand(\emb{x}', \emb{y})).$
\end{enumerate}
The vector $\emb{e}=\phi(C)$ is scored as $\mathrm{NIR}(\phi(C),\mathbf{a})=\sigma(f_{\phi_{\mathrm{last}}}(\emb{e},\mathbf{a}))$, where $\phi_{\mathrm{last}}=\phi_\nand$ for this concept. Because $\rho$ delegates atomic concepts to Pellet, the Compositional model tends to reproduce symbolic retrieval on complete knowledge bases but inherits Pellet's sensitivity to missing assertions.

\section{Experiments}
\label{sec:experiments}
\subsection{Experimental Setup}
We conducted instance retrieval experiments on 8 datasets to evaluate the performance of our approach: Semantic Bible~\cite{kouagou2023neural}, Animals, Carcinogenesis, Lymphography, Mutagenesis, Nctrer, Suramin~\cite{westphal2019sml}, and Family~\cite{demir2023neuro}. These datasets span a diverse range of domains, such as familial relationships, chemical compound interactions, disease diagnosis, and zoological taxonomies.

Three sets of experiments are considered. The first set concerns instance retrieval on complete knowledge bases (i.e., original datasets), where all our neural instance retrievers use \textsc{DeCal} embeddings; symbolic baselines and ground-truth labels are obtained with Pellet~\cite{sirin2007pellet}. The second set of experiments uses subsets of the original knowledge bases where ABox assertions are randomly deleted. Every deleted assertion either relates an individual to a concept (i.e.\ of the form $C(a)$) or represents a relationship between two individuals. A hyperparameter $\alpha\in\{0.25,0.5,0.75\}$ specifies the fraction of ABox assertions retained; the fraction removed is $1-\alpha$. As such, the constructed subsets of knowledge bases lack critical information for determining the true sets of instances of given class expressions, and we use them to assess the effectiveness of our proposed approach. The third experiment retrains GRU and Transformer with TransE~\cite{bordes2013translating} embeddings---a simpler, lower-expressiveness alternative---on complete knowledge bases (Section~\ref{sec:transe-results}).

\begin{table}[htb]
\centering
\caption{Evaluation datasets. $|N_C|$, $|N_R|$, and $|N_I|$ are the numbers of named concepts, named roles, and individuals, respectively.}
\label{tab:datasets}
\resizebox{0.99\textwidth}{!}{
\begin{tabular}{lcccccccc}
  \toprule
  & Animals & Carcinogenesis & Family & Lymphography & Mutagenesis & Nctrer & Semantic Bible & Suramin \\
  \midrule
  $|N_C|$ & 39 & 142 & 18 & 49 & 86 & 37 & 48 & 46 \\
  $|N_R|$ & 2 & 4 & 4 & 0 & 5 & 9 & 29 & 3 \\
  $|N_I|$ & 20 & 22372 & 202 & 148 & 14145 & 10209 & 724 & 2979 \\
  \midrule
   Train & 2983 & 20877 & 3484 & 3368 & 10556  & 4231 & 16000 & 2946\\
   Test & 746 & 5220 & 872 & 842 & 2640 & 1058 & 4000 & 737\\
  \bottomrule
\end{tabular}
}
\end{table}
In Table~\ref{tab:datasets}, we present the dataset statistics. Example~\ref{ex:animals-kb} illustrates the structure of the Animals benchmark; the other datasets follow the same DL-Learner format (TBox axioms plus an ABox of typed individuals and role assertions).

\begin{example}[Animals benchmark]
\label{ex:animals-kb}
The Animals knowledge base comprises a TBox with 39 named concepts, two object roles (\texttt{habitat}, \texttt{hasCovering}), and one datatype role (\texttt{hasLegs}), together with an ABox of 20 individuals. Each individual is typed with a single atomic concept in the ABox (e.g., \texttt{Eel(eel01)}); further properties are specified in the TBox via concept definitions. Table~\ref{tab:animals-individuals} lists all individuals. At test time, $\mathrm{NIR}$ receives a complex \dl\ concept $C$ and an individual $a$, and returns a score $p(C(a))$. Retrieval quality is measured by comparing the set of individuals predicted above a threshold against the ground-truth instance set computed by Pellet~\cite{sirin2007pellet} on the complete knowledge base.
\end{example}

\begin{table}[htb]
\centering
\small
\caption{Individuals in the Animals benchmark. The prefix \texttt{animals\#} denotes the DL-Learner namespace.}
\label{tab:animals-individuals}
\begin{tabular}{ll|ll}
\toprule
\textbf{Individual} & \textbf{Type} & \textbf{Individual} & \textbf{Type} \\
\midrule
\texttt{animals\#bat01} & Bat & \texttt{animals\#lizard01} & Lizard \\
\texttt{animals\#boy01} & Boy & \texttt{animals\#ostrich01} & Ostrich \\
\texttt{animals\#cat01} & Cat & \texttt{animals\#penguin01} & Penguin \\
\texttt{animals\#croco01} & Crocodile & \texttt{animals\#platypus01} & Platypus \\
\texttt{animals\#dog01} & Dog & \texttt{animals\#shark01} & Shark \\
\texttt{animals\#dolphin01} & Dolphin & \texttt{animals\#snake01} & Snake \\
\texttt{animals\#dragon01} & Dragon & \texttt{animals\#trex01} & T-Rex \\
\texttt{animals\#eagle01} & Eagle & \texttt{animals\#trout01} & Trout \\
\texttt{animals\#eel01} & Eel & \texttt{animals\#turtle01} & Turtle \\
\texttt{animals\#girl01} & Girl & \texttt{animals\#herring01} & Herring \\
\bottomrule
\end{tabular}
\end{table}

Training and test data are generated using a deterministic algorithm. The algorithm iterates through atomic concepts and role names, computes negated concepts and inverse roles, and constructs all description logic expressions in \dl with a length\footnote{The length of a concept is defined in \cite{kouagou2022learning}} of at most 15 (experimental choice) to prevent an excessive number of expressions. Also, note that we discard class expressions with empty instance sets, which leaves us with expressions of length at most 9 (see Figure~\ref{fig:expressionlengths}). The training sets are then further split into two subsets: 90\% for training and 10\% for validation. We used \textsc{DeCal}~\cite{kamdem2024embedding} to compute embeddings for individuals and roles. Full training hyperparameters are listed in Appendix~\ref{app:details} (Table~\ref{tab:hyperparams}).

\subsection{Evaluation Metrics and Hardware}
To evaluate the effectiveness of our approach in retrieving instances across different datasets, we employ two standard metrics: the Jaccard similarity (Jac) and the $\text{F}_1$ score. These metrics compare the set of retrieved instances, denoted by $\hat{y}$, against the ground truth set $y$. They offer a measure of how closely the predicted instance sets match ground truths, which are obtained with Pellet~\cite{sirin2007pellet}. These two metrics are defined as follows:
\begin{align}
\text{Jac}(\hat{y}, y) &=
\begin{cases}
\frac{|\hat{y} \cap y|}{|\hat{y} \cup y|} & \text{if } y \neq \emptyset \text{ or } \hat{y} \neq \emptyset, \\
1 & \text{otherwise},
\end{cases} \ \
\text{F}_1(\hat{y}, y) &=
\begin{cases}
\frac{2 \cdot |\hat{y} \cap y|}{|\hat{y}| + |y|} & \text{if } y \neq \emptyset \text{ or } \hat{y} \neq \emptyset, \\
1 & \text{otherwise}.
\end{cases}
\end{align}
In addition to retrieval accuracy, we also assess the computational efficiency of our approach, namely the retrieval runtime. All experiments were run on a Ubuntu 22.04 server equipped with a single NVIDIA H100-40C with 40GB VRAM and 16 Intel Xeon Platinum 8462Y+ CPUs @ 2.8 GHz. Our neural instance retrievers are lightweight, consuming only 1.2 to 3 GB of GPU VRAM during training with a batch size of 256.

\subsection{Main Results and Discussion}
Table~\ref{tab:merged-results} reports retrieval accuracy and runtime on complete knowledge bases (bold and underlined entries mark the best and second-best scores per row). No single architecture dominates every dataset: the Transformer attains the highest Jaccard on Carcinogenesis, Family, and Lymphography, the Compositional model leads on Animals, Nctrer, Semantic Bible, and Suramin, and GRU is best on Mutagenesis. Across architectures, Jaccard w.r.t.\ Pellet~\cite{sirin2007pellet} ranges from $69.1\%$ (LSTM on Suramin) to $99.4\%$ (Compositional on Animals), indicating that neural retrievers can approximate exact symbolic retrieval on complete knowledge bases.

\begin{figure}[ht]
\centering
\begin{tikzpicture}
\begin{axis}[
    hide axis,
    xmin=0, xmax=0.8,
    ymin=0, ymax=1,
    legend columns=5,
    legend cell align={left},
    legend style={
        draw=none,
        font=\small,
        column sep=8pt,
        /tikz/every even column/.append style={column sep=1em},
    },
]
\addlegendimage{blue!40, thick, mark=*, mark size=2pt}
\addlegendentry{Compositional}

\addlegendimage{red!40, mark=square*, mark size=2pt}
\addlegendentry{GRU}

\addlegendimage{cyan!40, mark=triangle*, mark size=2.5pt}
\addlegendentry{LSTM}

\addlegendimage{orange!40, mark=pentagon*, mark size=2.5pt}
\addlegendentry{Transformer}

\end{axis}
\end{tikzpicture}
\begin{tikzpicture}
\begin{groupplot}[
    group style={
        group size=2 by 2,
        horizontal sep=1.5cm,
        vertical sep=1.5cm,
    },
    width=0.45\textwidth,
    height=0.30\textwidth,
    ymin=0.7, ymax=1.0,
    xmin=1.5, xmax=9.5,
    xtick={2,3,4,5,6,7,8,9},
    ytick={0.7,0.8,0.9,1.0},
    ylabel={Avg. Jaccard},
    xlabel={Class Expression Length},
    ymajorgrids=true,
    grid style=dashed,
    legend style={
        at={(0.5,-0.25)},
        anchor=north,
        legend columns=4,
        /tikz/every even column/.append style={column sep=0.5cm}
    },
]

\nextgroupplot
\addplot+[ybar,fill=blue!40] coordinates {
    (2,0.9277) (3,0.8915) (4,0.9435) (5,0.9684)
    (6,0.9754) (7,0.9408) (8,0.9549) (9,0.9650)
};
\addplot+[only marks,mark=*,mark size=2.5pt,gray,mark options={fill=blue!40}] coordinates {
    (2,0.9277) (3,0.8915) (4,0.9435) (5,0.9684)
    (6,0.9754) (7,0.9408) (8,0.9549) (9,0.9650)
};

\nextgroupplot
\addplot+[ybar,fill=red!40] coordinates {
    (2,0.8734) (3,0.8764) (4,0.8906) (5,0.9272)
    (6,0.8701) (7,0.8771) (8,0.9043) (9,0.9091)
};
\addplot+[only marks,mark=square*,mark size=2.5pt,gray,mark options={fill=red!40}] coordinates {
    (2,0.8734) (3,0.8764) (4,0.8906) (5,0.9272)
    (6,0.8701) (7,0.8771) (8,0.9043) (9,0.9091)
};

\nextgroupplot
\addplot+[ybar,fill=cyan!40] coordinates {
    (2,0.8430) (3,0.7456) (4,0.8946) (5,0.9366)
    (6,0.8616) (7,0.8422) (8,0.8728) (9,0.8702)
};
\addplot+[only marks,mark=triangle*,mark size=2.5pt,gray,mark options={fill=cyan!40}] coordinates {
    (2,0.8430) (3,0.7456) (4,0.8946) (5,0.9366)
    (6,0.8616) (7,0.8422) (8,0.8728) (9,0.8702)
};

\nextgroupplot
\addplot+[ybar,fill=orange!40] coordinates {
    (2,0.8967) (3,0.9609) (4,0.8611) (5,0.9326)
    (6,0.9108) (7,0.9034) (8,0.8698) (9,0.9072)
};
\addplot+[only marks,mark=pentagon*,mark size=2.5pt,gray,mark options={fill=orange!40}] coordinates {
    (2,0.8967) (3,0.9609) (4,0.8611) (5,0.9326)
    (6,0.9108) (7,0.9034) (8,0.8698) (9,0.9072)
};
\end{groupplot}
\end{tikzpicture}
\caption{Average Jaccard scores by class expression length for different model architectures. Markers (disc, square, triangle, pentagon) indicate the actual performance for each architecture.}
\label{fig:expressionlengths}
\end{figure}
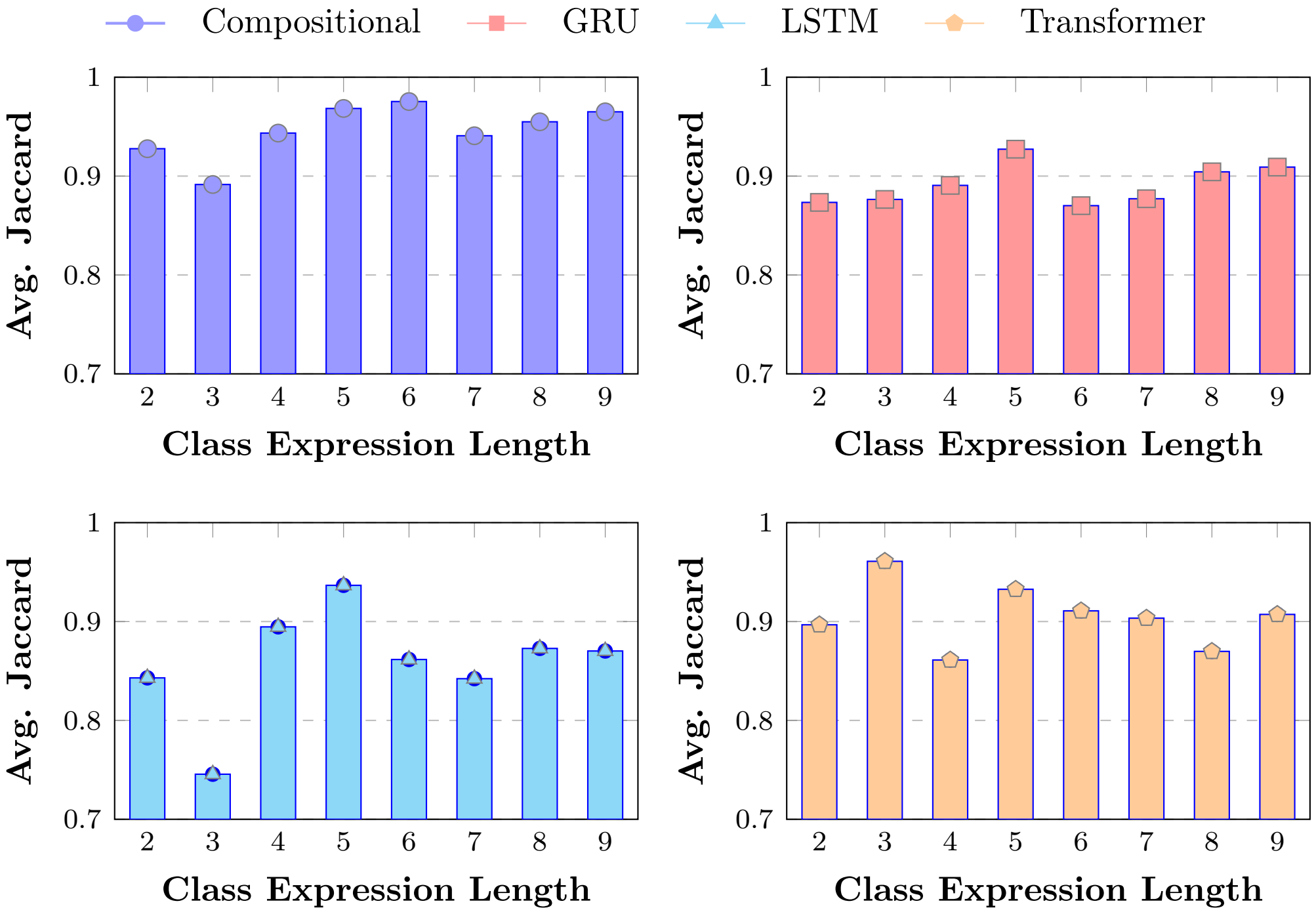

Figure~\ref{fig:expressionlengths} shows performance by concept length. Tokenization-based models are more sensitive to length and surface syntax, whereas the Compositional model remains stable on long expressions because each operator is handled by a fixed sub-network. This stability comes at a cost: by encoding atomic concepts through Pellet, the Compositional model behaves like a differentiable surrogate of symbolic retrieval and is less robust when assertions are removed (Section~\ref{sec:incomplete-results}).

\begin{table}[ht!]
\centering
\setlength{\tabcolsep}{2pt}
\caption{Retrieval results across all datasets. For each dataset, we report Jaccard, $F_1$-score, and relative $\Delta_{\text{Time}}$ (\%). The asterisk (*) indicates that the runtime difference between the respective model and Pellet is statistically significant (Wilcoxon signed-rank test, $p<0.05$). $\Delta_{\text{Time}}$ is $100\times(T_{\text{Pellet}}-T_{\text{model}})/T_{\text{Pellet}}$; values outside $[-1000,1000]$ are shown as \texttt{<-1000} or \texttt{>1000}. Hence, positive $\Delta_{\text{Time}}$ indicates that an approach is faster than Pellet. Empty entries for EBR indicate that it did not complete within 96 hours.}
\label{tab:merged-results}
\resizebox{0.9\linewidth}{!}{
\begin{tabular}{ l c c  c  c  c  c  c }
\toprule
\textbf{Dataset} && \textbf{Metric} & \textbf{Compositional} & \textbf{Transformer} & \textbf{GRU} & \textbf{LSTM} & \textbf{EBR} \\
\midrule
\multirow{4}{*}{Animals} && Jaccard & \textbf{0.994} & \underline{0.989} & 0.981 & 0.841 & 0.198 \\
                         && $F_1$-score & \textbf{0.995} & \underline{0.992} & 0.986 & 0.854 & 0.309 \\
                         && $\Delta_{\text{Time}}$ (\%) & -100* & \underline{-75}* & \textbf{-50}* & -150* & \texttt{<-1000}* \\
                         
\midrule
\multirow{4}{*}{Carcinogenesis} && Jaccard & 0.871 & \textbf{0.937} & \underline{0.894} & 0.802 & -- \\
                         && $F_1$-score & 0.917 & \textbf{0.952} & \underline{0.933} & 0.871 & -- \\
                         && $\Delta_{\text{Time}}$ (\%) & -78.2* & -5.5* & \textbf{33.8}* & \underline{-5.3}* & -- \\
\midrule
\multirow{4}{*}{Family}  && Jaccard & \underline{0.909} & \textbf{0.939} & 0.785 & 0.698 & 0.401 \\
                         && $F_1$-score & \underline{0.942} & \textbf{0.959} & 0.856 & 0.787 & 0.525 \\
                         && $\Delta_{\text{Time}}$ (\%) & -36.8* & \underline{5.3}* & \textbf{29.4}* & 0* & \texttt{<-1000}* \\
\midrule
\multirow{4}{*}{Lymphography}   && Jaccard & 0.921 & \textbf{0.986} & 0.793 & 0.691 & \underline{0.983} \\
                         && $F_1$-score & 0.951 & \textbf{0.991} & 0.860 & 0.781 & \underline{0.987} \\
                         && $\Delta_{\text{Time}}$ (\%) & \underline{-114.3}* & -200* & \textbf{-100}* & -233.3* & -666.7* \\
\midrule
\multirow{4}{*}{Mutagenesis} && Jaccard & 0.888 & \underline{0.897} & \textbf{0.919} & 0.825 & -- \\
                         && $F_1$-score & \underline{0.921} & 0.917 & \textbf{0.941} & 0.875 & -- \\
                         && $\Delta_{\text{Time}}$ (\%) & 40.9* & \underline{66.4}* & \textbf{73.5}* & 62.3* & -- \\
\midrule
\multirow{4}{*}{Nctrer}  && Jaccard & \textbf{0.972} & \underline{0.912} & 0.905 & 0.845 & 0.003 \\
                         && $F_1$-score & \textbf{0.980} & 0.925 & \underline{0.927} & 0.872 & 0.005 \\
                         && $\Delta_{\text{Time}}$ (\%) & 44.3* & \underline{80.8}* & \textbf{87.6}* & 80* & \texttt{<-1000}* \\
\midrule
\multirow{4}{*}{Semantic Bible}    && Jaccard & \textbf{0.978} & \underline{0.924} & 0.904 & 0.888 & 0.043 \\
                         && $F_1$-score & \textbf{0.982} & \underline{0.934} & 0.923 & 0.910 & 0.060 \\
                         && $\Delta_{\text{Time}}$ (\%) & -44.7* & 59.6* & \textbf{74.5}* & \underline{61.7}* & -936.2* \\
\midrule
\multirow{4}{*}{Suramin} && Jaccard & \textbf{0.947} & \underline{0.875} & 0.770 & 0.691 & 0.409 \\
                         && $F_1$-score & \textbf{0.962} & \underline{0.890} & 0.829 & 0.761 & 0.483 \\
                         && $\Delta_{\text{Time}}$ (\%) & 42.9 & 63.2* & \textbf{77.7}* & \underline{65.2}* & \texttt{<-1000}* \\
\bottomrule
\end{tabular}
}
\end{table}

In comparison to the baseline EBR, we can observe that our neural instance retrievers are clearly better on most datasets except for the Lymphography dataset, where EBR achieves a slightly higher Jaccard similarity than the Compositional, GRU and LSTM architectures. Additionally, the performance of EBR doesn't generalize well across datasets, as it achieves a Jaccard similarity of $98.3\%$ on the Lymphography dataset, but only $0.3\%$ on the Nctrer dataset. The poor performance of EBR is likely due to its high dependency on the quality of the representations learned by the embedding models to perform reasoning tasks, which can be challenging when dealing with complex concepts and large datasets. On the other hand, our neural retrievers can generalize better across different datasets, leading to more consistent performance. In Table \ref{tab:merged-results}, we can also see that EBR is computationally intensive on large datasets, as it fails to complete within 96 hours on the Mutagenesis and Carcinogenesis datasets, while our neural retrievers completed in less than 3 minutes on these datasets.  

Overall, tokenization-based encoders offer the best balance of accuracy and robustness for CEL pipelines: the Transformer is the most accurate on Carcinogenesis, Family, and Lymphography, and GRU offers the fastest inference. The Compositional model is competitive on complete knowledge bases and is preferable when new DL constructors must be added via modular sub-networks rather than retraining a sequence model. Beyond these aggregate metrics, every $\mathrm{NIR}$ variant returns a graded membership score rather than a binary yes/no decision. We illustrate this behavior on the Animals test set with the Transformer, our strongest tokenization-based retriever on several benchmarks (see \texttt{Inference.ipynb}\footnote{\url{https://github.com/fosterreproducibleresearch/NIR/blob/main/Inference.ipynb}}):
\begin{enumerate}
    \item[$\ast$]$\mathrm{NIR}$(\texttt{$\neg$Penguin $\sqcap\ \forall\ $hasCovering.Feathers, animals\#eagle01) = $0.999$},
    \item[$\ast$]$\mathrm{NIR}$(\texttt{$\neg$Eel $\sqcap$ $\neg$Bird, animals\#eel01) = $1.684\times 10^{-6}$}.
\end{enumerate}
These scores match our expectations: \emph{an eagle is not a penguin and is covered with feathers, and an eel is an instance of the class Eel, which is disjoint with $\neg$Eel}. Tokenization-based encoders can nonetheless be sensitive to surface syntax. The two cases below show that removing parentheses from an otherwise equivalent expression changes Transformer predictions substantially:
\begin{enumerate}
    \item[$\ast$]$\mathrm{NIR}$(\texttt{($\neg$Eel), animals\#eel01) = $0.288$},
    \item[$\ast$]$\mathrm{NIR}$(\texttt{$\neg$Eel, animals\#eel01) = $0.713$}.
\end{enumerate}

Regarding runtimes, our neural instance retrievers are highly competitive, especially on large datasets. Relative speedups (see $\Delta_{\text{Time}}$ in Table~\ref{tab:merged-results}) show that NIR is often substantially faster than Pellet on Carcinogenesis, Family, Mutagenesis, Nctrer, Semantic Bible, and Suramin, while Pellet remains faster on the smallest benchmarks (Animals and Lymphography). This is due to the fact that neural networks are generally fast at inference time since their prediction mechanisms are mostly based on efficient matrix operations, which can further be accelerated on GPUs. Wilcoxon signed-rank tests confirm these differences are statistically significant ($p<0.05$) in most cases, except for the Compositional model on Suramin. Among our models, GRU attains the largest relative speedups on most medium-to-large datasets, followed by the Transformer, LSTM, and Compositional architectures. EBR is comparatively much slower than our neural retrievers and Pellet (Table~\ref{tab:merged-results}).

Most importantly, our approach allows to predict the likelihood that a given individual is an instance of a concept, which is not supported by existing instance retrieval engines. This can be particularly useful when dealing with incomplete knowledge bases. In this case, a user may set a threshold $\gamma$ on output probability scores to decide which new assertions to accept and add to an original knowledge base. For example, the assertion $C(\texttt{Cake\_2})$, where $C=\exists\ \texttt{hasIngredient.Sugar}$, can be added to a reference knowledge base if our approach predicts a score higher than the given threshold, i.e., when our approach predicts a score higher than a set threshold.
Our approach hence lies between the closed and open-world assumptions~\cite{drummond2006open}, where the truth value of assertions is parameterized by the learned model weights and $\gamma$. For incomplete knowledge bases, we recommend a tokenization-based $\mathrm{NIR}$---in our experiments, the Transformer---over the Compositional variant.

\subsection{Results on Incomplete Knowledge Bases}
\label{sec:incomplete-results}
As mentioned in the experimental setup section, we evaluate $\mathrm{NIR}$ on incomplete knowledge bases where random ABox assertions are removed. Figure~\ref{fig:combined-incompleteness} reports average Jaccard similarity for all eight datasets at retention levels $\alpha\in\{0.25,0.5,0.75\}$; the point $1-\alpha{=}0$ uses the complete-knowledge-base scores from Table~\ref{tab:merged-results}. Tokenization-based encoders---especially the Transformer---remain competitive with or outperform Pellet under incompleteness on several benchmarks (e.g., Semantic Bible, Nctrer, and Suramin), while Pellet and the Compositional model degrade as more assertions are removed. On Family, Lymphography, and Carcinogenesis, all approaches decline more sharply, which is consistent with smaller or sparser ABoxes (Table~\ref{tab:datasets}). Across datasets, the Compositional curves track Pellet closely, confirming that this encoder is the nearest neural surrogate of symbolic retrieval and therefore less robust when assertions are missing. Tokenization-based $\mathrm{NIR}$ should therefore be preferred when knowledge bases may be incomplete.

\newsavebox{\NIRAnimalsIncPlot}
\savebox{\NIRAnimalsIncPlot}{
\begin{minipage}{0.24\textwidth}
\centering
\begin{tikzpicture}
\begin{axis}[
    width=\linewidth,
    height=2.8cm,
    ylabel={Avg. Jaccard},
    xmin=0, xmax=0.8,
    ymin=0.0, ymax=1.0,
    grid=both,
    grid style={dashed,gray!40},
    xlabel={$1{-}\alpha$},
]
\addplot[blue, thick, mark=*, mark size=2pt] coordinates { (0.00, 0.989) (0.25, 0.880) (0.50, 0.706) (0.75, 0.912) };
\addplot[orange, thick, dashed, mark=square*, mark size=1.8pt] coordinates { (0.00, 0.981) (0.25, 0.740) (0.50, 0.734) (0.75, 0.681) };
\addplot[green!60!black, thick, dash dot, mark=diamond*, mark size=1.8pt] coordinates { (0.00, 0.841) (0.25, 0.681) (0.50, 0.681) (0.75, 0.681) };
\addplot[red, thick, dotted, mark=triangle*, mark size=2pt] coordinates { (0.00, 0.994) (0.25, 0.992) (0.50, 0.993) (0.75, 0.994) };
\addplot[cyan, thick, mark=triangle*, mark options={rotate=180}, mark size=2pt] coordinates { (0.00, 1.00) (0.25, 1.00) (0.50, 1.00) (0.75, 1.00) };
\end{axis}
\end{tikzpicture}
\end{minipage}%
}

\newsavebox{\NIRCarcinoIncPlot}
\savebox{\NIRCarcinoIncPlot}{
\begin{minipage}{0.24\textwidth}
\centering
\begin{tikzpicture}
\begin{axis}[
    width=\linewidth,
    height=2.8cm,
    xmin=0, xmax=0.8,
    ymin=0.0, ymax=1.0,
    grid=both,
    grid style={dashed,gray!40},
    xlabel={$1{-}\alpha$},
]
\addplot[blue, thick, mark=*, mark size=2pt] coordinates { (0.00, 0.937) (0.25, 0.659) (0.50, 0.670) (0.75, 0.650) };
\addplot[orange, thick, dashed, mark=square*, mark size=1.8pt] coordinates { (0.00, 0.894) (0.25, 0.404) (0.50, 0.387) (0.75, 0.395) };
\addplot[green!60!black, thick, dash dot, mark=diamond*, mark size=1.8pt] coordinates { (0.00, 0.802) (0.25, 0.279) (0.50, 0.271) (0.75, 0.277) };
\addplot[red, thick, dotted, mark=triangle*, mark size=2pt] coordinates { (0.00, 0.871) (0.25, 0.589) (0.50, 0.543) (0.75, 0.529) };
\addplot[cyan, thick, mark=triangle*, mark options={rotate=180}, mark size=2pt] coordinates { (0.00, 1.00) (0.25, 0.992) (0.50, 0.712) (0.75, 0.716) };
\end{axis}
\end{tikzpicture}
\end{minipage}%
}

\newsavebox{\NIRFamilyIncPlot}
\savebox{\NIRFamilyIncPlot}{
\begin{minipage}{0.24\textwidth}
\centering
\begin{tikzpicture}
\begin{axis}[
    width=\linewidth,
    height=2.8cm,
    xmin=0, xmax=0.8,
    ymin=0.0, ymax=1.0,
    grid=both,
    grid style={dashed,gray!40},
    xlabel={$1{-}\alpha$},
]
\addplot[blue, thick, mark=*, mark size=2pt] coordinates { (0.00, 0.939) (0.25, 0.715) (0.50, 0.551) (0.75, 0.304) };
\addplot[orange, thick, dashed, mark=square*, mark size=1.8pt] coordinates { (0.00, 0.785) (0.25, 0.545) (0.50, 0.440) (0.75, 0.250) };
\addplot[green!60!black, thick, dash dot, mark=diamond*, mark size=1.8pt] coordinates { (0.00, 0.698) (0.25, 0.511) (0.50, 0.424) (0.75, 0.244) };
\addplot[red, thick, dotted, mark=triangle*, mark size=2pt] coordinates { (0.00, 0.909) (0.25, 0.567) (0.50, 0.345) (0.75, 0.157) };
\addplot[cyan, thick, mark=triangle*, mark options={rotate=180}, mark size=2pt] coordinates { (0.00, 1.00) (0.25, 0.650) (0.50, 0.435) (0.75, 0.231) };
\end{axis}
\end{tikzpicture}
\end{minipage}%
}

\newsavebox{\NIRLymphIncPlot}
\savebox{\NIRLymphIncPlot}{
\begin{minipage}{0.24\textwidth}
\centering
\begin{tikzpicture}
\begin{axis}[
    width=\linewidth,
    height=2.8cm,
    xmin=0, xmax=0.8,
    ymin=0.0, ymax=1.0,
    grid=both,
    grid style={dashed,gray!40},
    xlabel={$1{-}\alpha$},
]
\addplot[blue, thick, mark=*, mark size=2pt] coordinates { (0.00, 0.986) (0.25, 0.980) (0.50, 0.981) (0.75, 0.980) };
\addplot[orange, thick, dashed, mark=square*, mark size=1.8pt] coordinates { (0.00, 0.793) (0.25, 0.779) (0.50, 0.791) (0.75, 0.788) };
\addplot[green!60!black, thick, dash dot, mark=diamond*, mark size=1.8pt] coordinates { (0.00, 0.691) (0.25, 0.456) (0.50, 0.405) (0.75, 0.604) };
\addplot[red, thick, dotted, mark=triangle*, mark size=2pt] coordinates { (0.00, 0.921) (0.25, 0.666) (0.50, 0.422) (0.75, 0.204) };
\addplot[cyan, thick, mark=triangle*, mark options={rotate=180}, mark size=2pt] coordinates { (0.00, 1.00) (0.25, 0.725) (0.50, 0.436) (0.75, 0.207) };
\end{axis}
\end{tikzpicture}
\end{minipage}%
}

\newsavebox{\NIRMutagenIncPlot}
\savebox{\NIRMutagenIncPlot}{
\begin{minipage}{0.24\textwidth}
\centering
\begin{tikzpicture}
\begin{axis}[
    width=\linewidth,
    height=2.8cm,
    ylabel={Avg. Jaccard},
    xmin=0, xmax=0.8,
    ymin=0.0, ymax=1.0,
    grid=both,
    grid style={dashed,gray!40},
    xlabel={$1{-}\alpha$},
]

\addplot[blue, thick, mark=*, mark size=2pt] coordinates { (0.00, 0.897) (0.25, 0.811) (0.50, 0.810) (0.75, 0.809) };
\addplot[orange, thick, dashed, mark=square*, mark size=1.8pt] coordinates { (0.00, 0.919) (0.25, 0.528) (0.50, 0.516) (0.75, 0.536) };
\addplot[green!60!black, thick, dash dot, mark=diamond*, mark size=1.8pt] coordinates { (0.00, 0.825) (0.25, 0.462) (0.50, 0.444) (0.75, 0.460) };
\addplot[red, thick, dotted, mark=triangle*, mark size=2pt] coordinates { (0.00, 0.888) (0.25, 0.711) (0.50, 0.620) (0.75, 0.634) };
\addplot[cyan, thick, mark=triangle*, mark options={rotate=180}, mark size=2pt] coordinates { (0.00, 1.00) (0.25, 0.984) (0.50, 0.715) (0.75, 0.698) };
\end{axis}
\end{tikzpicture}
\end{minipage}%
}

\newsavebox{\NIRNctrerIncPlot}
\savebox{\NIRNctrerIncPlot}{
\begin{minipage}{0.24\textwidth}
\centering
\begin{tikzpicture}
\begin{axis}[
    width=\linewidth,
    height=2.8cm,
    xmin=0, xmax=0.8,
    ymin=0.0, ymax=1.0,
    grid=both,
    grid style={dashed,gray!40},
    xlabel={$1{-}\alpha$},
]
\addplot[blue, thick, mark=*, mark size=2pt] coordinates { (0.00, 0.912) (0.25, 0.972) (0.50, 0.976) (0.75, 0.975) };
\addplot[orange, thick, dashed, mark=square*, mark size=1.8pt] coordinates { (0.00, 0.905) (0.25, 0.917) (0.50, 0.915) (0.75, 0.919) };
\addplot[green!60!black, thick, dash dot, mark=diamond*, mark size=1.8pt] coordinates { (0.00, 0.845) (0.25, 0.895) (0.50, 0.883) (0.75, 0.868) };
\addplot[red, thick, dotted, mark=triangle*, mark size=2pt] coordinates { (0.00, 0.972) (0.25, 0.880) (0.50, 0.849) (0.75, 0.822) };
\addplot[cyan, thick, mark=triangle*, mark options={rotate=180}, mark size=2pt] coordinates { (0.00, 1.00) (0.25, 0.930) (0.50, 0.862) (0.75, 0.818) };
\end{axis}
\end{tikzpicture}
\end{minipage}%
}

\newsavebox{\NIRSembIncPlot}
\savebox{\NIRSembIncPlot}{
\begin{minipage}{0.24\textwidth}
\centering
\begin{tikzpicture}
\begin{axis}[
    width=\linewidth,
    height=2.8cm,
    xmin=0, xmax=0.8,
    ymin=0.0, ymax=1.0,
    grid=both,
    grid style={dashed,gray!40},
    xlabel={$1{-}\alpha$},
]
\addplot[blue, thick, mark=*, mark size=2pt] coordinates { (0.00, 0.924) (0.25, 0.976) (0.50, 0.974) (0.75, 0.975) };
\addplot[orange, thick, dashed, mark=square*, mark size=1.8pt] coordinates { (0.00, 0.904) (0.25, 0.932) (0.50, 0.931) (0.75, 0.942) };
\addplot[green!60!black, thick, dash dot, mark=diamond*, mark size=1.8pt] coordinates { (0.00, 0.888) (0.25, 0.878) (0.50, 0.891) (0.75, 0.902) };
\addplot[red, thick, dotted, mark=triangle*, mark size=2pt] coordinates { (0.00, 0.978) (0.25, 0.956) (0.50, 0.875) (0.75, 0.847) };
\addplot[cyan, thick, mark=triangle*, mark options={rotate=180}, mark size=2pt] coordinates { (0.00, 1.00) (0.25, 0.970) (0.50, 0.875) (0.75, 0.844) };
\end{axis}
\end{tikzpicture}
\end{minipage}%
}

\newsavebox{\NIRSuraminIncPlot}
\savebox{\NIRSuraminIncPlot}{
\begin{minipage}{0.24\textwidth}
\centering
\begin{tikzpicture}
\begin{axis}[
    width=\linewidth,
    height=2.8cm,
    xmin=0, xmax=0.8,
    ymin=0.0, ymax=1.0,
    grid=both,
    grid style={dashed,gray!40},
    xlabel={$1{-}\alpha$},
]
\addplot[blue, thick, mark=*, mark size=2pt] coordinates { (0.00, 0.875) (0.25, 0.914) (0.50, 0.905) (0.75, 0.909) };
\addplot[orange, thick, dashed, mark=square*, mark size=1.8pt] coordinates { (0.00, 0.770) (0.25, 0.666) (0.50, 0.639) (0.75, 0.622) };
\addplot[green!60!black, thick, dash dot, mark=diamond*, mark size=1.8pt] coordinates { (0.00, 0.691) (0.25, 0.605) (0.50, 0.597) (0.75, 0.622) };
\addplot[red, thick, dotted, mark=triangle*, mark size=2pt] coordinates { (0.00, 0.947) (0.25, 0.798) (0.50, 0.646) (0.75, 0.607) };
\addplot[cyan, thick, mark=triangle*, mark options={rotate=180}, mark size=2pt] coordinates { (0.00, 1.00) (0.25, 0.969) (0.50, 0.710) (0.75, 0.634) };
\end{axis}
\end{tikzpicture}
\end{minipage}%
}

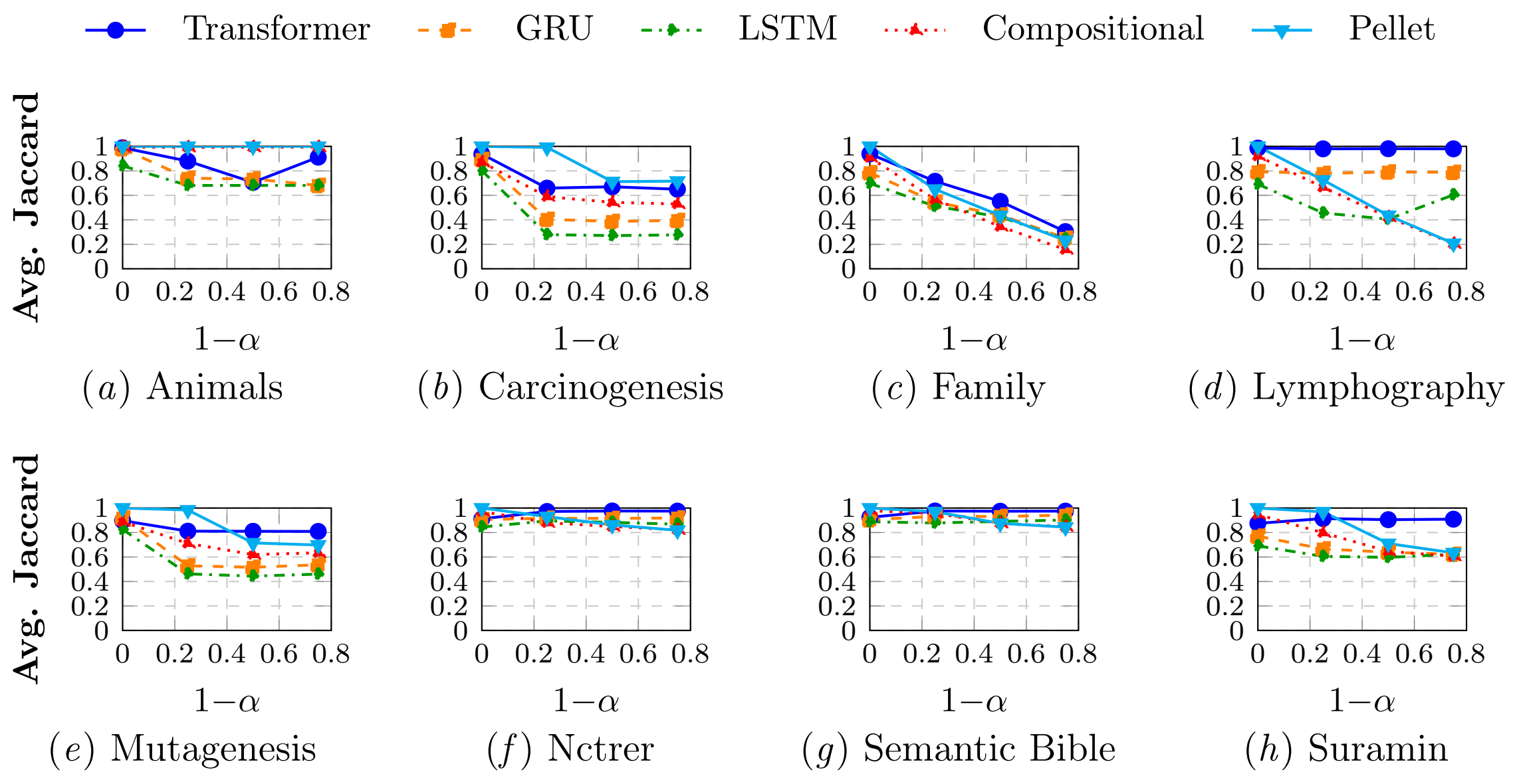
\begin{figure*}[t]
\floatconts
{fig:combined-incompleteness}
{\caption{Average Jaccard for instance retrievers at different incompleteness levels across all eight datasets. $\alpha$ is the fraction of ABox assertions retained ($\alpha\in\{0.25,0.5,0.75\}$); the horizontal axis shows $1-\alpha$, the fraction removed. At $1-\alpha{=}0$, scores are taken from the complete knowledge bases (Table~\ref{tab:merged-results}). Ground-truth instance sets are computed with Pellet on the original complete knowledge bases.}}
{%
\begin{tikzpicture}
\begin{axis}[
    hide axis,
    xmin=0, xmax=0.8,
    ymin=0, ymax=1,
    legend columns=5,
    legend cell align={left},
    legend style={
        draw=none,
        font=\small,
        column sep=8pt,
        /tikz/every even column/.append style={column sep=1em},
    },
]
\addlegendimage{blue, thick, mark=*, mark size=2pt}
\addlegendentry{Transformer}
\addlegendimage{orange, thick, dashed, mark=square*, mark size=1.8pt}
\addlegendentry{GRU}
\addlegendimage{green!60!black, thick, dash dot, mark=diamond*, mark size=1.8pt}
\addlegendentry{LSTM}
\addlegendimage{red, thick, dotted, mark=triangle*, mark size=2pt}
\addlegendentry{Compositional}
\addlegendimage{cyan, thick, mark=triangle*, mark options={rotate=180}, mark size=2pt}
\addlegendentry{Pellet}
\end{axis}
\end{tikzpicture}

\vspace{0.4em}
\subfigure[Animals]{\usebox{\NIRAnimalsIncPlot}}\hfill
\subfigure[Carcinogenesis]{\usebox{\NIRCarcinoIncPlot}}\hfill
\subfigure[Family]{\usebox{\NIRFamilyIncPlot}}\hfill
\subfigure[Lymphography]{\usebox{\NIRLymphIncPlot}}\\[0.6em]
\subfigure[Mutagenesis]{\usebox{\NIRMutagenIncPlot}}\hfill
\subfigure[Nctrer]{\usebox{\NIRNctrerIncPlot}}\hfill
\subfigure[Semantic Bible]{\usebox{\NIRSembIncPlot}}\hfill
\subfigure[Suramin]{\usebox{\NIRSuraminIncPlot}}%
}
\end{figure*}

\subsection{Results with TransE Embeddings}
\label{sec:transe-results}
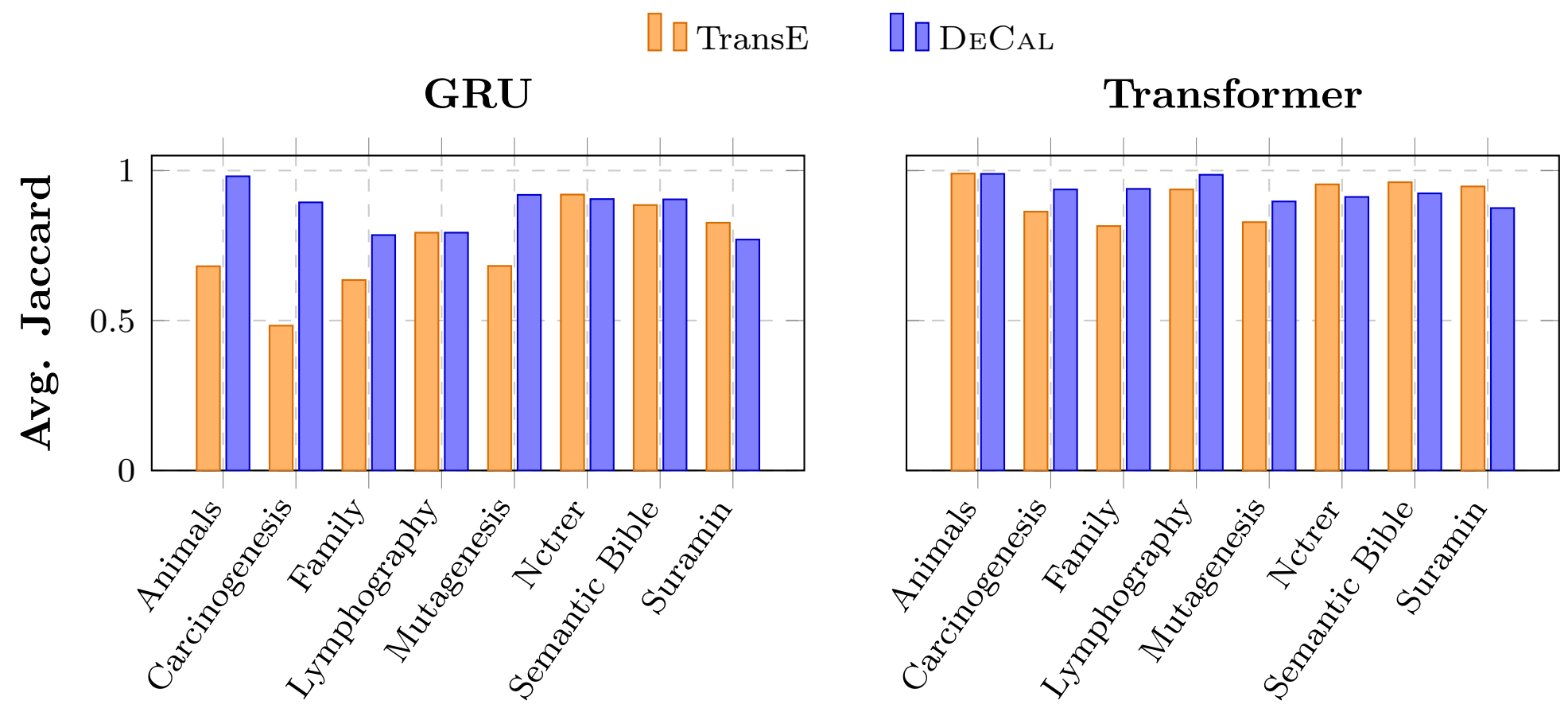
\begin{figure}[H]
\centering
\begin{tikzpicture}
\begin{groupplot}[
    group style={
        group size=2 by 1,
        horizontal sep=0.85cm,
        y descriptions at=edge left,
    },
    width=0.46\linewidth,
    height=4.2cm,
    ymin=0,
    ymax=1.05,
    ylabel={Avg. Jaccard},
    symbolic x coords={Animals,Carcinogenesis,Family,Lymphography,Mutagenesis,Nctrer,{Semantic Bible},Suramin},
    xtick=data,
    xticklabel style={rotate=55, anchor=east, font=\scriptsize},
    grid=both,
    grid style={dashed,gray!40},
    enlarge x limits=0.14,
]
\nextgroupplot[
    title={\small\textbf{GRU}},
    title style={at={(0.5,1.02)}, anchor=south},
    ybar,
    bar width=5.5pt,
    legend to name=transeLegend,
    legend style={
        draw=none,
        legend columns=2,
        font=\scriptsize,
        /tikz/every even column/.append style={column sep=0.6cm},
    },
]
\addplot[fill=orange!60, draw=orange!85!black, bar shift=-3.5pt] coordinates {
    (Animals, 0.681) (Carcinogenesis, 0.483) (Family, 0.635) (Lymphography, 0.793)
    (Mutagenesis, 0.682) (Nctrer, 0.920) ({Semantic Bible}, 0.885) (Suramin, 0.826)
};
\addlegendentry{TransE}
\addplot[fill=blue!50, draw=blue!80!black, bar shift=3.5pt] coordinates {
    (Animals, 0.981) (Carcinogenesis, 0.894) (Family, 0.785) (Lymphography, 0.793)
    (Mutagenesis, 0.919) (Nctrer, 0.905) ({Semantic Bible}, 0.904) (Suramin, 0.770)
};
\addlegendentry{\textsc{DeCal}}

\nextgroupplot[
    title={\small\textbf{Transformer}},
    title style={at={(0.5,1.02)}, anchor=south},
    ylabel={},
    ybar,
    bar width=5.5pt,
]
\addplot[fill=orange!60, draw=orange!85!black, bar shift=-3.5pt] coordinates {
    (Animals, 0.990) (Carcinogenesis, 0.863) (Family, 0.815) (Lymphography, 0.937)
    (Mutagenesis, 0.828) (Nctrer, 0.954) ({Semantic Bible}, 0.961) (Suramin, 0.947)
};
\addplot[fill=blue!50, draw=blue!80!black, bar shift=3.5pt] coordinates {
    (Animals, 0.989) (Carcinogenesis, 0.937) (Family, 0.939) (Lymphography, 0.986)
    (Mutagenesis, 0.897) (Nctrer, 0.912) ({Semantic Bible}, 0.924) (Suramin, 0.875)
};
\end{groupplot}
\node[anchor=south] at ($(group c1r1.north west)!0.5!(group c2r1.north east)+(0,0.55cm)$) {%
    \pgfplotslegendfromname{transeLegend}%
};
\end{tikzpicture}
\caption{Average Jaccard on complete knowledge bases for GRU and Transformer with TransE and \textsc{DeCal} embeddings.}
\label{fig:transe-results}
\end{figure}

To assess whether $\mathrm{NIR}$ depends on a specific embedding method, we retrained the GRU and Transformer retrievers with TransE embeddings~\cite{bordes2013translating} on complete knowledge bases, keeping all other hyperparameters identical to the main \textsc{DeCal} experiment (Table~\ref{tab:merged-results}). Figure~\ref{fig:transe-results} compares average Jaccard for both embedding methods.
On most datasets, both architectures remain competitive when paired with the simpler TransE embeddings (Figure~\ref{fig:transe-results}). The main exception is Carcinogenesis, where GRU drops from $89.4\%$ (\textsc{DeCal}) to $48.3\%$ (TransE), suggesting that large, sparse knowledge bases may benefit from more expressive embeddings. The Transformer is less sensitive to the embedding choice and attains higher Jaccard with TransE than with \textsc{DeCal} on four datasets: Nctrer ($95.4\%$ vs.\ $91.2\%$), Semantic Bible ($96.1\%$ vs.\ $92.4\%$), Suramin ($94.6\%$ vs.\ $87.5\%$), and Animals ($99.0\%$ vs.\ $98.9\%$). Overall, medium-sized benchmarks are relatively insensitive to the embedding choice, whereas scale and expressiveness matter more for the largest datasets.

\FloatBarrier
\section{Conclusion}
\label{sec:conclusion}
In this paper, we introduced $\mathrm{NIR}$, a family of neural instance retrievers for \dl, and compared tokenization-based encoders with a Compositional design on eight benchmarks. All variants approximate Pellet closely on complete knowledge bases and are substantially faster at inference. Our main empirical finding is architectural: tokenization-based models---especially the Transformer---generalize better on large and incomplete knowledge bases, whereas the Compositional model tracks symbolic retrieval on complete knowledge bases and supports modular extension to new DL constructors. For CEL pipelines, we recommend a tokenization-based $\mathrm{NIR}$ by default, reserving the Compositional design for settings that require operator-level extensibility. Probabilistic outputs further allow graded assertion acceptance and trade-offs between closed- and open-world reasoning via a user-defined threshold $\gamma$.


\bibliography{main}

\appendix

\section{Experimental Details}
\label{app:details}

Table~\ref{tab:hyperparams} summarizes the shared training configuration ($d{=}256$, hidden 128, dropout 0.1, $\gamma{=}0.5$; vocabulary 512, max length 128; Transformer: 3 layers / 4 heads).

\begin{table}[H]
\centering
\caption{Shared training hyperparameters for all $\mathrm{NIR}$ variants.}
\label{tab:hyperparams}
\small
\begin{tabular}{@{}ll@{}}
\toprule
\textbf{Setting} & \textbf{Value} \\
\midrule
Optimizer & Adam, learning rate $3.5\times10^{-4}$ \\
LR schedule & Cosine annealing \\
Epochs & 400 (no early stopping) \\
Batch size & 256 \\
Train/validation split & 90\% / 10\% \\
Embedding dimension $d$ & 256 \\
Hidden size & 128 \\
Dropout & 0.1 \\
Classification threshold $\gamma$ & 0.5 \\
Max sequence length & 128 \\
Vocabulary size & 512 \\
RNN layers (GRU/LSTM) & 3 \\
Transformer layers / heads & 3 / 4 \\
Hardware & NVIDIA H100-40GB, Ubuntu 22.04 \\
\bottomrule
\end{tabular}
\end{table}

\end{document}